\documentclass[letterpaper, 10 pt, conference]{ieeeconf}  

\IEEEoverridecommandlockouts                              

\usepackage{graphicx, subfigure} 
\usepackage{amsmath} 
\usepackage{color}
\usepackage{rotating}
\usepackage{hyperref}

\begin{document}

\title{\LARGE \bf
Incremental Real-Time Multibody VSLAM with Trajectory Optimization Using Stereo Camera 
}

\author{N Dinesh Reddy$^{1,2}$, Iman Abbasnejad$^{2,3,4}$, Sheetal Reddy$^{1}$, Amit Kumar Mondal$^{5}$ and Vindhya Devalla$^{5}$
\thanks{$^{1}$ International Institute of Information Technology, Hyderabad, India}%
\thanks{$^{2}$ Max Planck Institute For Intelligent Systems, T{\"u}bingen, Germany}%
\thanks{$^{3}$ The Robotics Institute, Pittsburgh, PA, USA }%
\thanks{$^{4}$ Queensland University of Technology, Brisbane, QLD, Australia}%
\thanks{$^{5}$ University of Petroleum and Energy Studies, Dehradun, India}%
}

\maketitle
\thispagestyle{empty}
\pagestyle{empty}

\begin{abstract}
Real-time outdoor navigation in highly dynamic environments is an crucial problem. The recent literature on real-time static SLAM don't scale up to dynamic outdoor environments. Most of these methods assume moving objects as outliers or discard the information provided by them. We propose an algorithm to jointly infer the camera trajectory and the moving object trajectory simultaneously. In this paper, we perform a sparse scene flow based motion segmentation using a stereo camera. The segmented objects motion models are used for accurate localization of the camera trajectory as well as the moving objects. We exploit the relationship between moving objects for improving the accuracy of the poses. We formulate the poses as a factor graph incorporating all the constraints. We achieve exact incremental solution by solving a full nonlinear optimization problem in real time. The evaluation is performed on the challenging KITTI dataset with multiple moving cars.Our method outperforms the previous baselines in outdoor navigation.

\end{abstract}

\section{INTRODUCTION}
Outdoor navigation in dynamic environments is a challenging task for autonomous driving assistance systems (ADAS). It has wide applications in various areas like collision avoidance, path planning and scene understanding. Inference of highly dynamic scenes accurately is crucial for such systems. A robot navigating in such environments needs a fast and accurate localization of the moving objects and their trajectories. In the past decade, a lot of literature is available for static SFM or SLAM~\cite{kerl13iros, engel14eccv} pipeline, where they utilize the static landmarks to build an accurate map and trajectory. These methods treat the moving objects as outliers. These methods fail when there are multiple moving objects in the scene like a congested urban scene due to wrong inlier fitting.\\

\begin{figure}[t!]
\centering
\includegraphics[width=60mm]{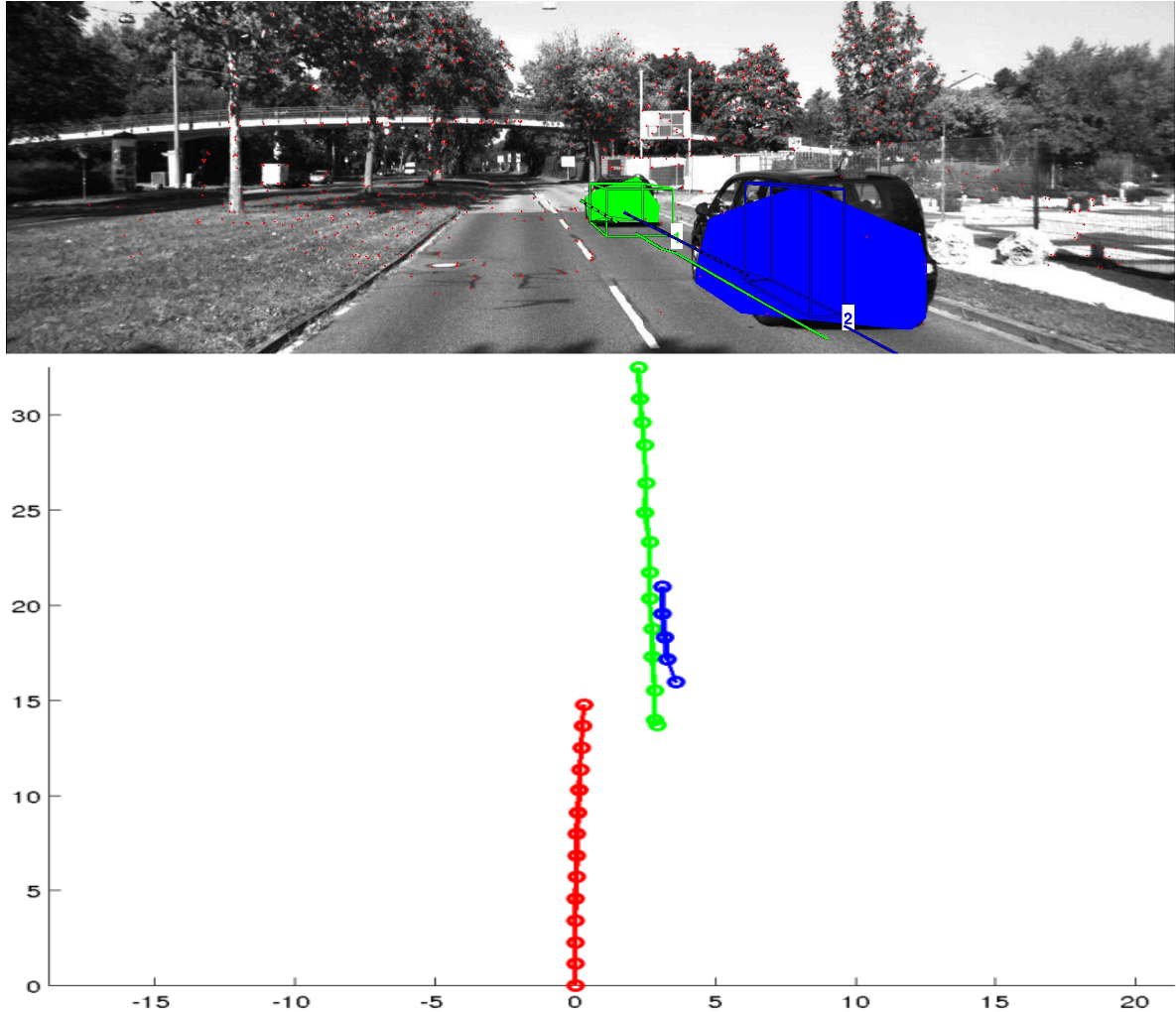}
\caption{
The top images depicts the real time segmentation of moving objects. The top view of the trajectories of moving objects and the camera odometry are depicted. The Red trajectory represents the camera odometry. The blue and green represents the moving car odometry.Best viewed in color.
}
\label{fig:first}
\end{figure}
Dynamic object segmentation and trajectory optimization is relatively new field of research with sparse literature. The few solutions to this problem present in the literature can be categorized into Decoupled and Joint Methods. Joint approach like \cite{roussos2012dense} use monocular cameras to jointly estimate the depth maps, do motion segmentation and motion estimation of multiple bodies. Decoupled approaches like \cite{DBLP:conf/cvpr/YuanM06,MulVSLAM_Abhijit_ICCV2011} have a sequential pipeline where they segment motion and independently reconstruct the moving and static scenes. Our approach is a real-time incremental approach, and differs from the other methods due to the simultaneous optimization of multiple moving cars. The algorithm is easily scalable to multiple cars and highly traffic scenarios.

Our approach emphasis on a real time approach for moving object trajectory optimization. We obtain our real time motion detection and segmentation from the sparse motion segmentation algorithm~\cite{lenz}. The moving object trajectories are initialized using the triangulation of the moving objects in the current frame and then transforming the trajectories with respect to the world coordinate frame. These trajectories are initialized as poses of a factor graph. The factor graph is included with additional constraints like the relationship between the camera motion and moving object. We also incorporate the motion model of the moving objects for more accurate localization. The main contribution in this paper is the optimization for the moving object trajectory and the camera trajectory simultaneously in real-time. We introduce the concept of anchor nodes for moving object trajectory estimation. The anchor nodes initialize each moving object as a new pose optimization problem and solves for the complete trajectory of the moving object.
\begin{figure*}[t!]
\centering
\includegraphics[width=165mm]{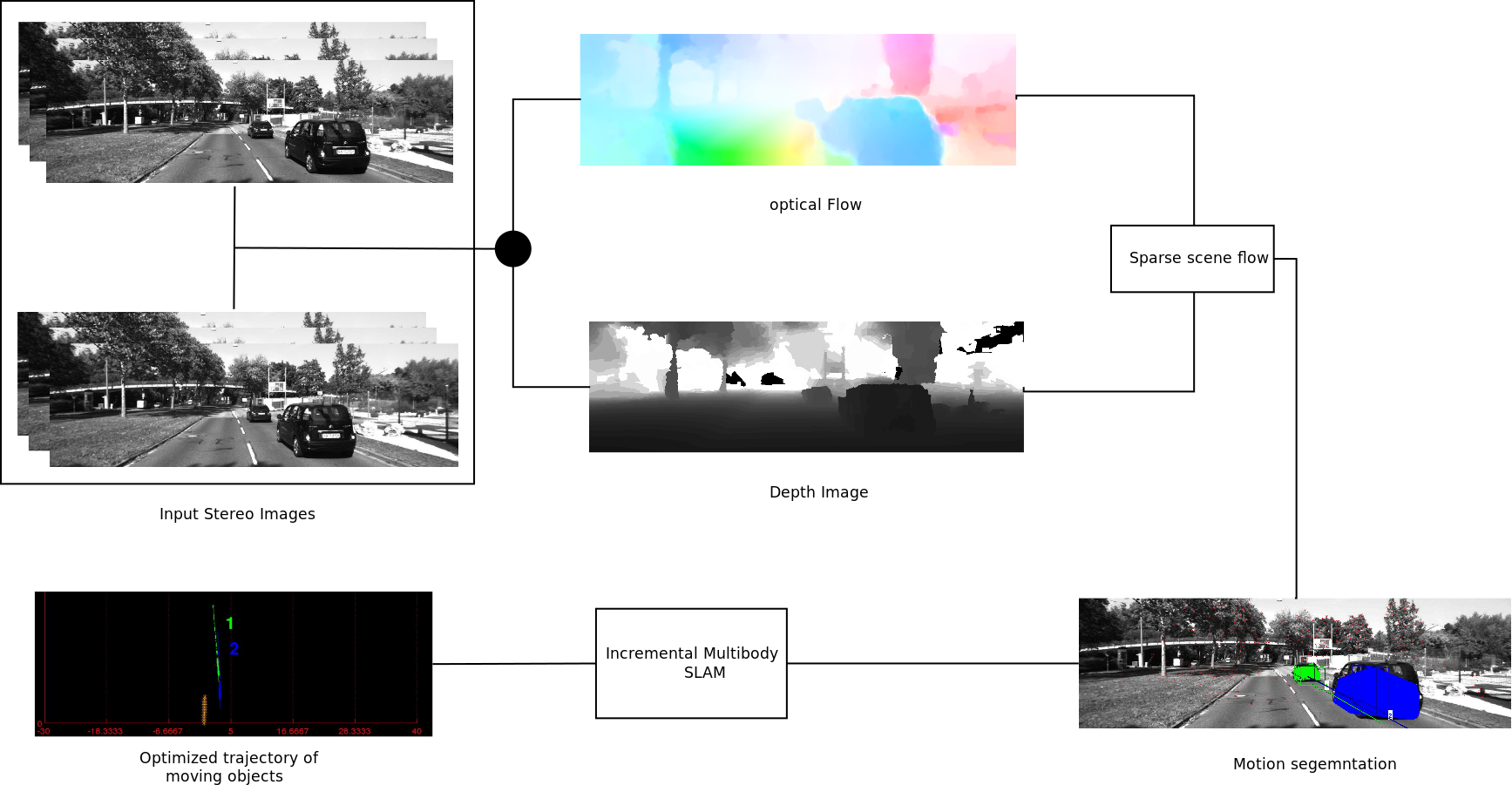}
\caption{Illustration of the proposed method.The system takes a sequence of rectified stereo images from the tracking dataset of KITTI (A).Our formulation computes the sparse scene flow (D) using disparity map(B) and optical flow(C).These are used to segment the multiple moving cars in the scene(E).For each segmented car, we estimate the trajectory separately and put them into a joint formulation. The optimized moving object trajectories are displayed in world coordinate frame. Best viewed in color.
}
\label{fig:pipeline}
\end{figure*}

\section{RELATED WORK}
A lot of research in the area of computer vision and robotics has been focused on motion segmentation and SLAM, but limited literature focuses on improving the localization of moving objects and their reconstruction using vision based feedback.
    Motion segmentation has been approached using geometric priors mostly from a video. General paradigm involves using geometric constraints \cite{Namdev2012,Romero-CanoN13}, reducing the model to affine to cluster the trajectories into subspaces~\cite{subspace2009} or semantic constraints~\cite{semanticmotion}.

A variety of approaches have been proposed to recover the Trajectories of static indoor and outdoor scenes. In contrast, here we propose a method that is able to extract accurate 3d information by reasoning jointly about static and dynamic scene elements as well as their complex inter-play using semantic information. Recent advances in static SFM involve adding semantic and geometric constraints to Bundle Adjustment \cite{sun3d}. Jianxiong et al. have shown results on indoor sequences with very high accuracy. Factor graphs\cite{Dellaert05rss} have shown considerable improvement in static robot localization. An incremental version of the smoothing and mapping \cite{Kaess11icra} has been shown a improvement in the Static incremental SLAM algorithms.

The moving object localization has recently been a widely researched area for Automated Driver Assistance Systems (ADAS). Song et al.~\cite{objectlocalization} use the  object detection and SFM cues for improving the 3D object localization. Dinesh et al.\cite{semanticloc} have used semantic constraints for accurate localization of moving objects. The accurate localization of the moving objects in dynamic environments helps in better understanding for outdoor navigation.

Our work closely resembles \cite{Kim10icra} in problem formulation. They exploit the relationship between two moving robots in the environment and solve the SLAM problem in an incremental formulation. We differ from the above method in terms of the constraints we exploit. We formulate our problem as a factor graph over moving object trajectories. This allows us to model smooth trajectories without employing hard geometric constraints. We also use these trajectories to fuse dynamic and static objects which can be used in robot navigation.

\section{OUR APPROACH}
We present a probabilistic formulation of the multi-body SLAM problem based on pose graphs. Pose graphs are a common solution for single robot localization and mapping, in which all current and past robot poses form a Markov chain connected by odometry measurements. We have implemented the incremental smoothing and mapping (iSAM) for optimizing the pose graph because it provides an efficient solution without need for approximations and allows effective access to the estimation uncertainties. The implementation exploits constraints between the different poses.

\subsection{Motion Detection and segmentation}
\label{sec:seg}
We consider a sequence of images from a stereo camera rig. Interest points are detected in two consecutive and rectified stereo images and checked for mutual consistency. The interest  points are generated using the SIFT feature detection algorithm. Each feature is matched with their stereo rig and the disparity is computed for each interest point. The 3d location of each Interest point is computed from the disparity. Since ,the mounting and pitching of the stereo rig is unknown, we detect the ground plane for better understanding of the environment. The ground plane computation gives a good prior for the computation of moving and stationary point clusters. All the cluster of points on the ground plane are segmented as stationary. These interest points are tracked over multiple frames and the scene flow is computed using finite difference approximation to yield derivatives. The scene flow stores the information of the motion of the interest point in the world frame.

A graph-like structure connecting all detected interest points in the image plane is generated using Delaunay triangulation. The resulting edges are removed according  to scene flow differences exceeding a certain threshold with respect to the uncertainty of the computed 3D position of every interest point. We have added additional geometric constraints for accurate segmentation as proposed in the~\cite{Romero-CanoN13}. We use the ground plane computed earlier for excluding the false positive solutions. The remaining connected components of the graph describe moving objects in the scene. Detected objects are tracked over time using a global nearest neighbor (GNN) approach. The GNN algorithm searches the closest object in distance from its location and tracks the object over multiple frames.

\begin{figure}[t!]
\centering
\includegraphics[width=70mm]{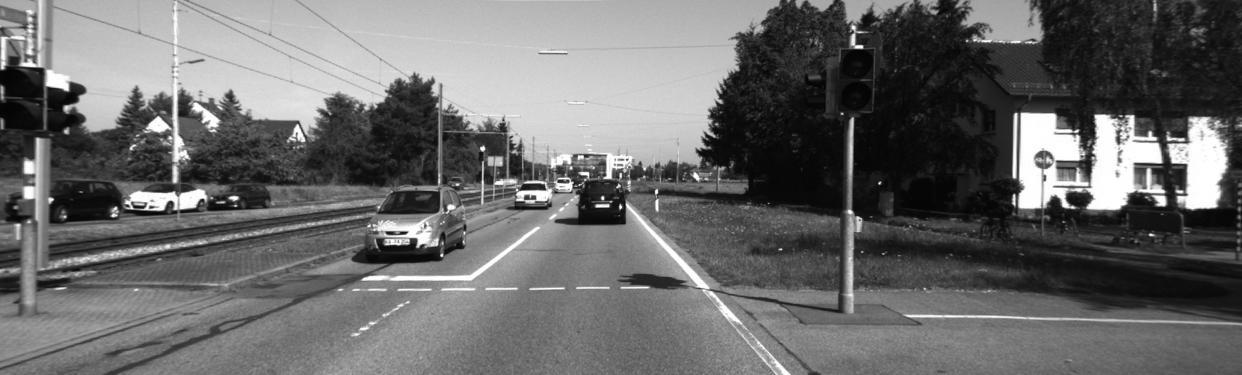}
\includegraphics[width=70mm]{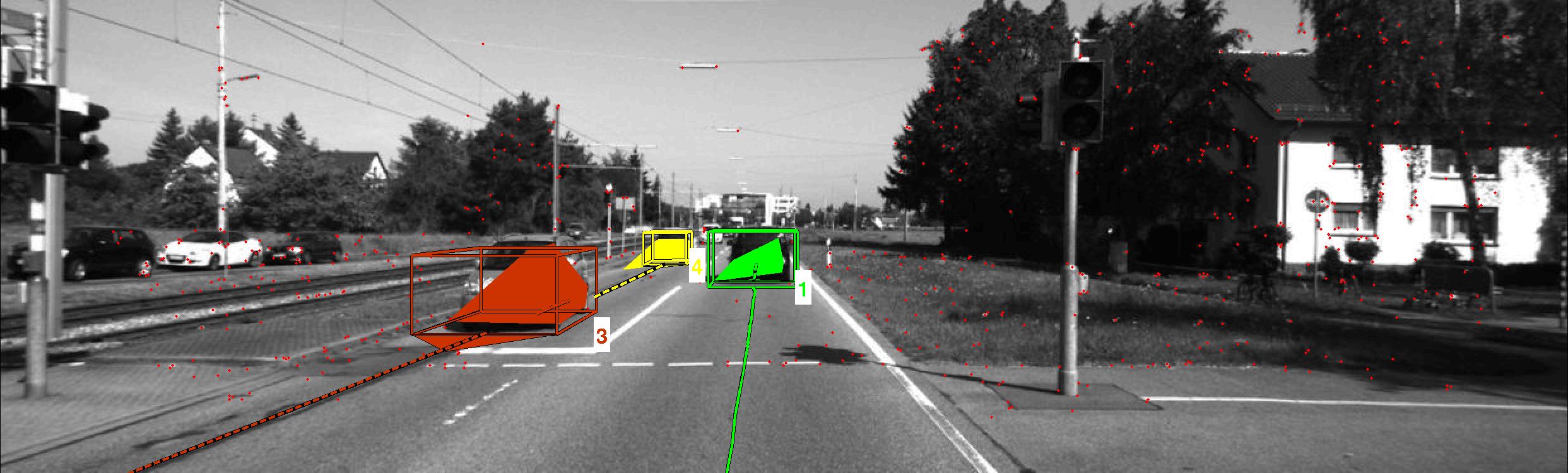}
\caption{The results for the motion segmentation algorithm on the KITTI2 sequence is depicted in the image.Each moving car is segmented and labelled with different color, the red point represent the stationary landmarks used for odometry computation. Best viewed in color.
}
\label{fig:motion_seg}
\end{figure}

\subsection{Moving object trajectories in global frame}

We first formulate the problem of the multibody mapping problem using one pose graph for each moving object trajectory. We show a typical moving object scenario with two moving objects in front of the camera in Fig.\ref{fig:first}. The pose variables are shown as coloured circles , and measurements as small black discs. In Fig.\ref{fig:first} each moving object is represented with different color. For $M$ moving objects, the trajectory of the moving object $m \in \lbrace 0....M-1 \rbrace $ in the scene is given by $N_m+1$ pose variables $\lbrace z_i^m \rbrace_{i=0}^{N_m}$. As each moving object trajectory is computed from the stereo triangulation, the trajectories by themselves are under-constrained. We fix the gauge freedom by introducing a prior $P^m$ for each trajectory $m$. Measurements between poses of a single trajectory are of two types. Where the successive poses are connected based on the camera readings, the other kind of measurements is the connection of arbitrary poses i.e readings of the poses between the cars from the depth information of the stereo cameras.

\begin{figure}[t!]
\centering
\includegraphics[width=60mm]{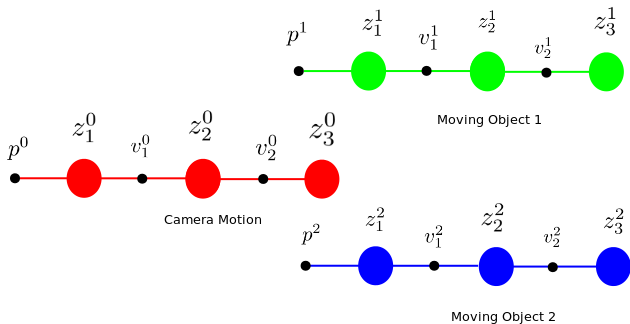}
\caption{
This depicts the formulation of the dynamic iSAM problem. The red represents the camera odometry, while the green and blue represent the moving objects predicted from the \ref{sec:seg}.
}
\label{fig:first}
\end{figure}

\begin{figure}[t!]
\centering
\includegraphics[width=60mm]{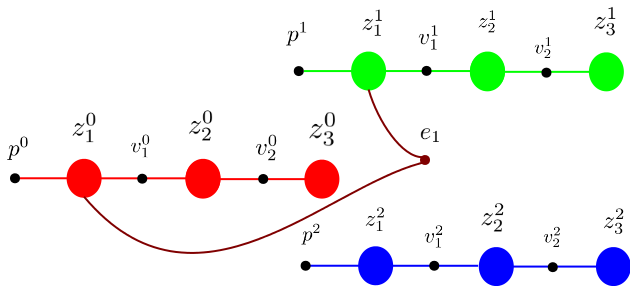}
\caption{
Here we depict the relationship between the moving object pose and the camera trajectory. The pose of the moving object and the camera is exploited by this constraint.
}
\label{fig:second}
\end{figure}

\begin{figure}[t!]
\centering
\includegraphics[width=60mm]{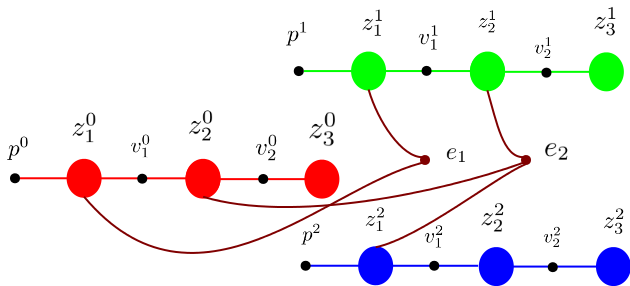}
\caption{ Here we depict the relationship between the moving object pose $m_1$ and another moving object $m_2$. This exploits the relationship between the moving object motions.
}
\label{fig:third}
\end{figure}

We have discussed about the moving car trajectory as an independent entity , We now introduce the relationships between the moving objects and the camera trajectories. An encounter $\bf e$ between the moving object and the camera odometry exploits the relationship between the camera motion and the moving object motion ,depicted in  Fig. \ref{fig:second}.Similarly, an encounter $\bf e $ between two moving objects $m_1$ and $m_2$ is a measurement that connects the two moving objects in a scene $z_i^{m_1}$ and $z_i^{m_2}$. An example is shown in the Fig. \ref{fig:third}, with the relation between the measurements and poses. Since we have a setup where the observations are the same time instance, all our measurements $\bf e $ connect poses taken at the same instance.

We take a probabilistic approach for estimating the moving object trajectory based on all the measurements.We formulate a joint probability for all the poses $Z= \lbrace z_i^m \rbrace_{i=0,m=0}^{N_m}$,measurements and priors $Y = \lbrace v_i^m \rbrace_{i=0}^{N_m} \cup \lbrace p^m \rbrace_{i=0}^{N_m}$  and L encounters $E= \lbrace e_i^m \rbrace_{i=0,m=0}^{N_m}$ :
\begin{equation}
\begin{split}
P(Z,Y,E) = \ \ \ \ \ \ \ \ \ \ \ \ \ \ \ \ \ \ \ \ \ \ \ \ \ \ \ \ \ \ \ \ \ \ \ \ \ \ \ \ \ \ \ \ \ \ \ \ \ \ \  \\  
\prod_{m=0}^{M-1} \Bigg(P(z_0^m|p_0^m) \prod_{i=1}^{N_m} P(z_i^m | z_{i-1}^m ,v_i)\Bigg) \prod_{j=1}^L P(z_{i_j}^{m_j} | z_{i'_j}^{m'_j},e_j)
\end{split}
\end{equation}
The number of encounters L is equal to to the number of frames in the sequence. The data association between the encounters ${i_j,i'_j,m_j,m'_j}$ is computed for each frame from the motion segmentation and localization algorithm as discussed in section \ref{sec:seg}. The noise is assumed to be gaussian as proposed in multiple SLAM systems.

\begin{equation}
z_i^m = f_i(z_{i-1}^m,v_i^m) + w_i^m
\end{equation}
This describes the moving object localization in the current frame of reference computed from the disparity computation. The $w_i^m$ is a normally distributed noise with covariance matrix $\xi_i^m $. We model the encounters within the moving objects and between the moving objects and the camera as :
\begin{equation}
z_{i_j}^{m_j} = h_j{z_{i'_j}^{m'_j},e_j}+k_j
\end{equation}

Similar to the successive poses interactions the noise for the measurement model of the moving objects is computed from the disparity computation. Here, $k_j$ is a normally distributed noise with covariance of $\Gamma_j$.

We formulate the problem as a maximum a posterior (MAP) estimate for the moving object trajectory. This leads to  the following nonlinear least squares problem:

\begin{equation}
\begin{split}
Z^* = arg \ min \Bigg \lbrace \sum_{r=0}^{M-1} \Big(|| p^m - z_0^{m}||^2_{\Sigma}  \\
 + \sum _{i=1}^{N_m} || f_i(z_{i-1}^m,v_i^m) + w_i^m ||^2_{\xi_i^m} \Big) \\
 + \sum_{j=1}^L \Big{|}\Big{|} h_j{z_{i'_j}^{m'_j},e_j}+z_{i_j}^{r_j} \Big{|}\Big{|}^2_{\Gamma_j} \Bigg \rbrace
 \end{split}
\end{equation}

Here $||a||_{\Sigma}^2 = a^T \Sigma^{-1}  a$ is the squared Mahalanobis distance with covariance matrix $\Sigma$.

We solve the non-linear least squares problem using the incremental smoothing and mapping (iSAM) algorithm. Since the error in object localization of the moving object is modelled as a gaussian and the constraint functions are nonlinear , nonlinear optimization methods are used. We can use the methods like gauss-newton, Levenberg-marquardt or the Powell's Dog-leg algorithm , which use succession of linear approximations to reach a minimum.All the components can be written in a standard least squares problem of the form :
\begin{equation}
\Theta^* = arg min_\Theta || A \Theta - b||^2
\end{equation}

where the vector $ \Theta \in R^n $ consists of all the moving object poses and the robot pose, where n is the number of variables. The matrix $A \in R^{mxn}$ is a large , but sparse measurement jacobian , with m the number of measurements, and $ b \in R^m$ is the right-hand side vector. We solve this using the method of QR factorization of the $A$ matrix.
\begin{equation}
A = Q \begin{bmatrix} H \\ 0 \end{bmatrix}
\end{equation}

where $H \in R^{nxn}$ is an $nxn$ upper triangulation matrix , $0$ is an $(m-n)xn$ zero matrix , $Q$ is an $mxn$ orthogonal matrix. The vector $b$ is modified accordingly during the QR decomposition to obtain $d \in R^n$. The solution is obtained by back substitution.
\begin{equation}
R \Theta = d
\end{equation}
To avoid refactoring an increasingly large measurement jacobian each time a new measurement is computed, we have followed the method of iSAM to update the new measurement rows. The key to efficiency is to keep the square root information matrix sparse, which requires choosing a suitable variable ordering. iSAM periodically reorders the variables according to some heuristic and performs a batch factorization that also includes relinearization of the measurement equations.

The initialization of the moving object trajectories is done using the stereo triangulation of all the points in the segmentation. The pose is initialized as the transformation between the set of 2d points. The motion detection algorithm is used for accurate prediction of the prior $p^m$, therefore we have very less gauss freedom and a very good initialization of the moving object trajectories.
\begin{table*}[t!]
\centering
\setlength\tabcolsep{1.5pt}
\begin{tabular}{|c|c|c|c|c|c|c|c|c|c|c|}
\hline
Car Num & P & \multicolumn{4}{c|}{\textbf{VISO2}} & \multicolumn{5}{c|}{\textbf{OUR APPROACH}} \tabularnewline

 &  & ATE R(m) & ATE M(m) & ATE Me(m) & ARE (deg)  & ATE R(m) & ATE M(m) & ATE Me(m) & ARE (deg) & {\bf RE(\%)} \tabularnewline
\hline
\hline
1 & 28 & 0.731 & 0.513 & 0.330 & 3.659 &  0.317 & 0.277 & 0.299 & 2.589 & {\bf 36.63} \tabularnewline
\hline
2 & 28 & 0.391 & 0.212 & 0.079 & 2.235 &  0.193 & 0.174 & 0.141 & 1.356 & {\bf 50.51} \tabularnewline
\hline
Cam & 28 & 0 & 0 & 0 & 0 & 0.10 & 0.09 & 0.08 & 1.238 &  \tabularnewline
\hline
\end{tabular}
\caption{Statistics of VISO2 and Our Approach for \textbf{KITTI 2} dataset.
\textbf{P} is \#poses. \textbf{ATE R} is
absolute trajectory error rmse, \textbf{ATE M} is absolute trajectory error mean,\textbf{ATE Me}
 is absolute trajectory error median, \textbf{ARE }is average rotation error and \textbf{RE} is relative pose error.
}
\label{tab:KITTI2_ate}
\end{table*}

\begin{table*}
\centering
\setlength\tabcolsep{1.5pt}

\begin{tabular}{|c|c|c|c|c|c|c|c|c|c|c|}
\hline
Car Num & P & \multicolumn{4}{c|}{\textbf{VISO2}} & \multicolumn{5}{c|}{\textbf{OUR APPROACH}} \tabularnewline
 &  & ATE R(m) & ATE M(m) & ATE Me(m) & ARE (deg)  & ATE R(m) & ATE M(m) & ATE Me(m) & ARE (deg) & {\bf RE(\%)}
\tabularnewline
\hline
\hline
1 & 212 & 2.371 & 2.222 & 2.009 & 5.272 &  2.146 & 1.822 & 1.307 & 3.594 & {\bf 9.4} \tabularnewline
\hline
2 & {\bf 9} & 0.900 & 0.744 & 0.733 & 3.685 &  0.519 & 0.446 & 0.436 & 2.976 & {\bf 42.33}  \tabularnewline
\hline
3 & {\bf 8} & 0.962 & 0.749 & 0.595 & 4.522 &  0.378 & 0.300 & 0.296 & 2.684 & {\bf 60.7}  \tabularnewline
\hline
4 & {\bf 8} & 0.381 & 0.324 & 0.307 & 2.354 &  0.259 & 0.217 & 0.214 & 1.343 & {\bf 32.02}  \tabularnewline
\hline
5 & {\bf 9} & 0.559 & 0.489 & 0.445 & 3.254 &  0.473 & 0.341 & 0.239 & 2.378 & {\bf 38.99}  \tabularnewline
\hline
Cam & 212 & 0 & 0 & 0 & 0 &  2.02 & 1.84 & 1.56 & 5.256 &   \%\tabularnewline
\hline
\end{tabular}
\caption{Statistics of VISO2 and Our Approach for \textbf{KITTI 1} dataset.
\textbf{P} is \#poses. \textbf{ATE R} is
absolute trajectory error rmse, \textbf{ATE M} is absolute trajectory error mean,\textbf{ATE Me}
 is absolute trajectory error median, \textbf{ARE }is average rotation error and \textbf{RE} is relative pose error.
 }
\label{tab:KITTI1_ate}
\end{table*}

\begin{figure*}[t!]
\begin{center}
\begin{tabular}{c c c c c c}
 & \bf Car 1 & \bf Car 2 & \bf Car 3  & \bf Car 4 & \bf Car 5\\
\begin{sideways}\bf \centering  \ \ \  \ \   VISO2 \end{sideways} & \includegraphics[width=30mm]{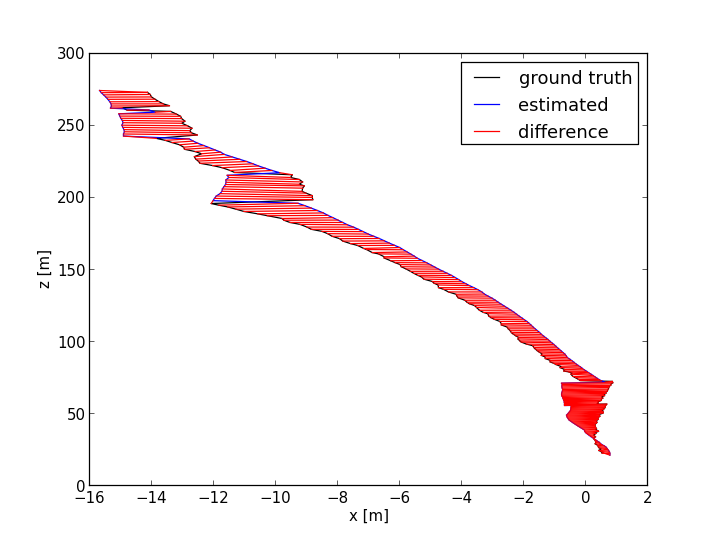} & \includegraphics[width=30mm]{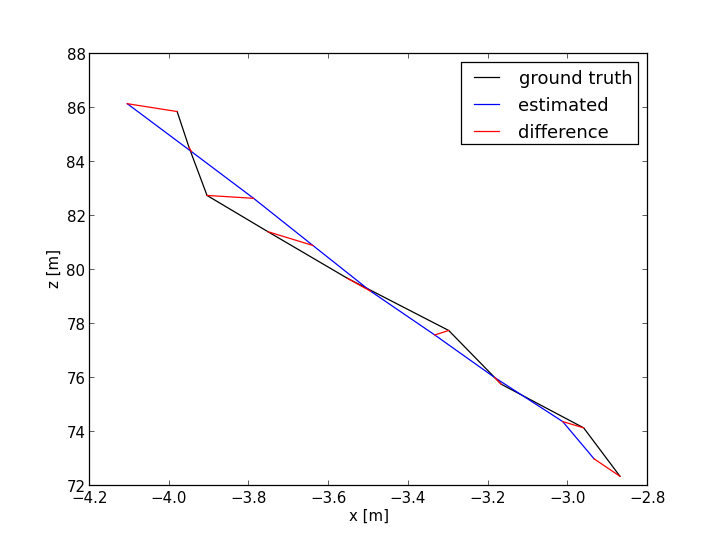} & \includegraphics[width=30mm]{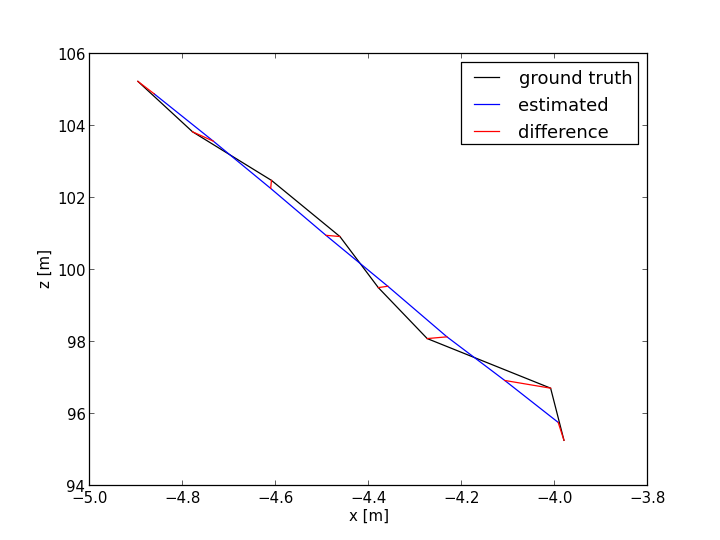} & \includegraphics[width=30mm]{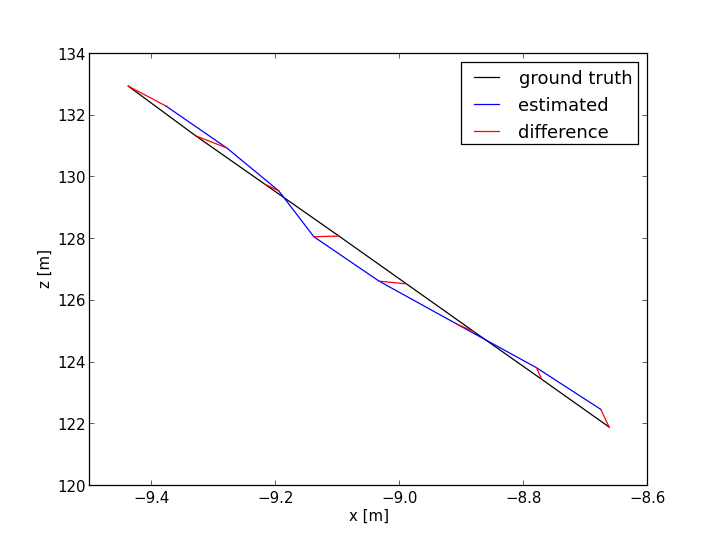} & \includegraphics[width=30mm]{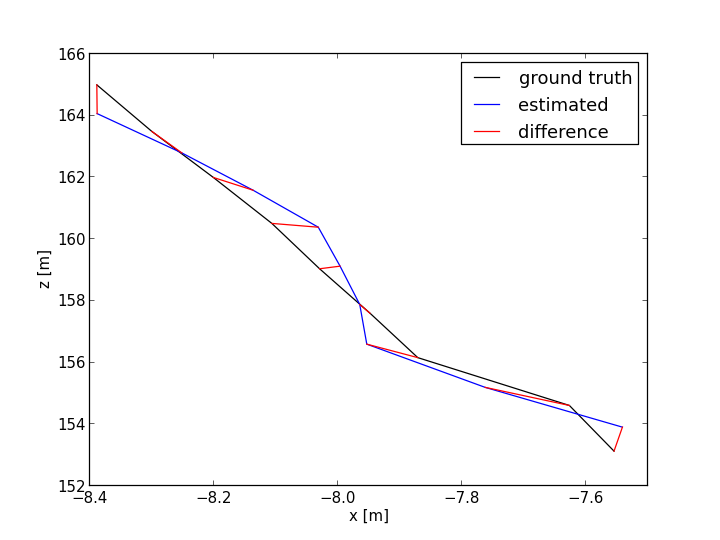} \\
\begin{sideways}\bf \centering  \ \ \ \  \    OURS\end{sideways} & \includegraphics[width=30mm]{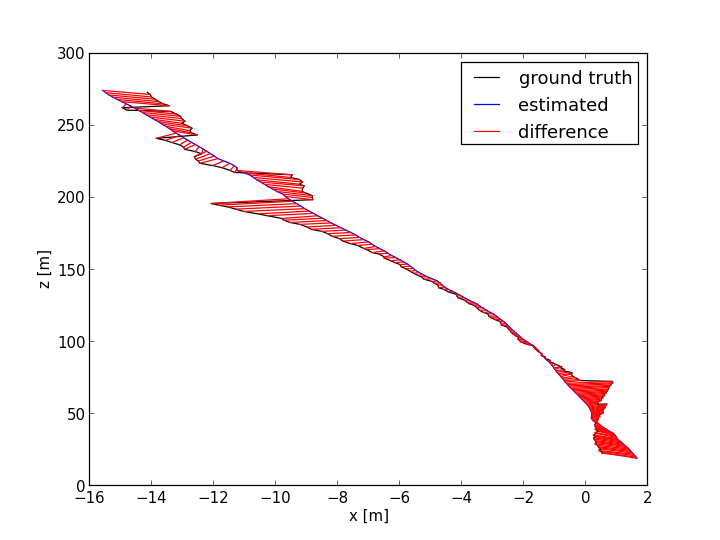} &
\includegraphics[width=30mm]{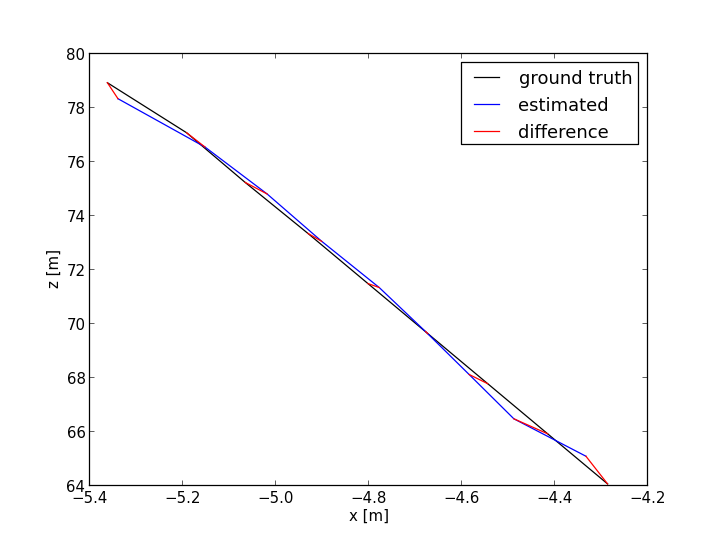} &
\includegraphics[width=30mm]{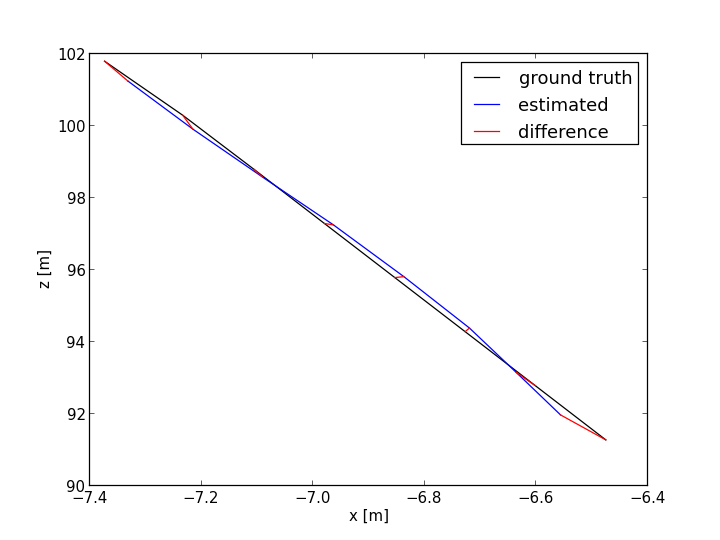} &
\includegraphics[width=30mm]{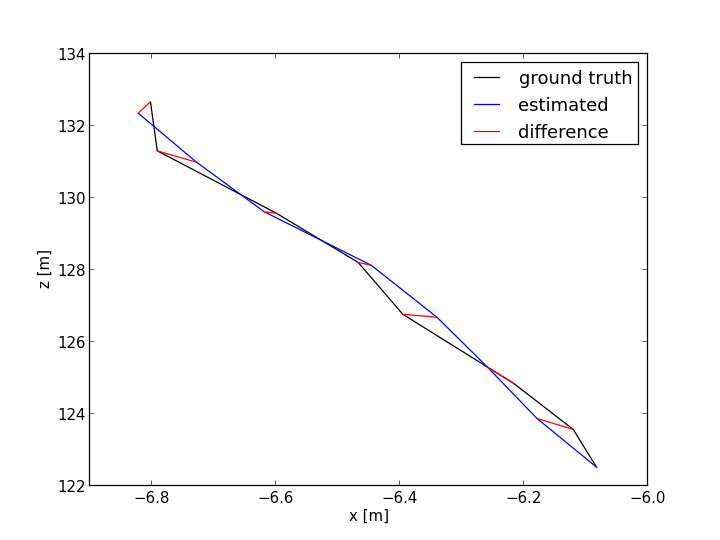} &
\includegraphics[width=30mm]{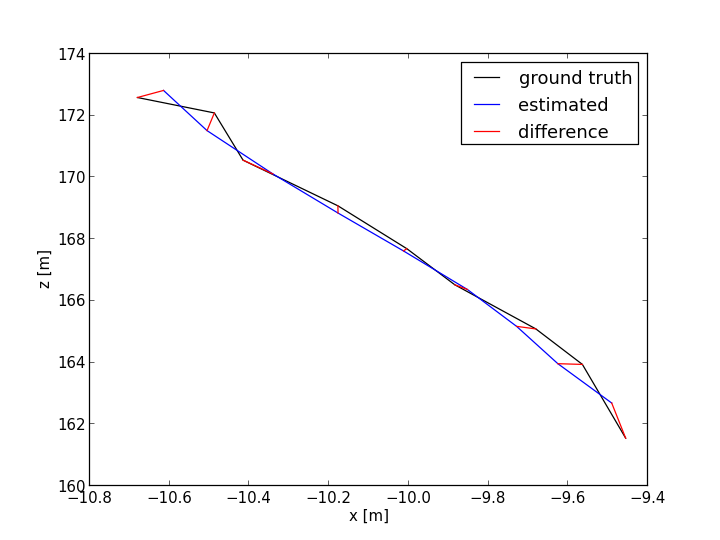}\\
\end{tabular}
\end{center}
\label{fig:figure_212}
 \caption{
Comparison plots for 5 moving cars in the KITTI 1. The black plot represents the ground truth trajectory of the moving car in the world frame. Blue plot represents the estimated trajectory of the moving car. Red lines represent the error in the estimate with respect to ground truth for the trajectories. The error comparison is computed between OUR method and VISO2.
}
\end{figure*}

\begin{figure}
\begin{center}
\begin{tabular}{c c c c}
 & \bf Car 1 & \bf Car 2  \\
\begin{sideways}\bf \centering  \ \ \ \ \ \   VISO2 \end{sideways} & \includegraphics[width=33mm]{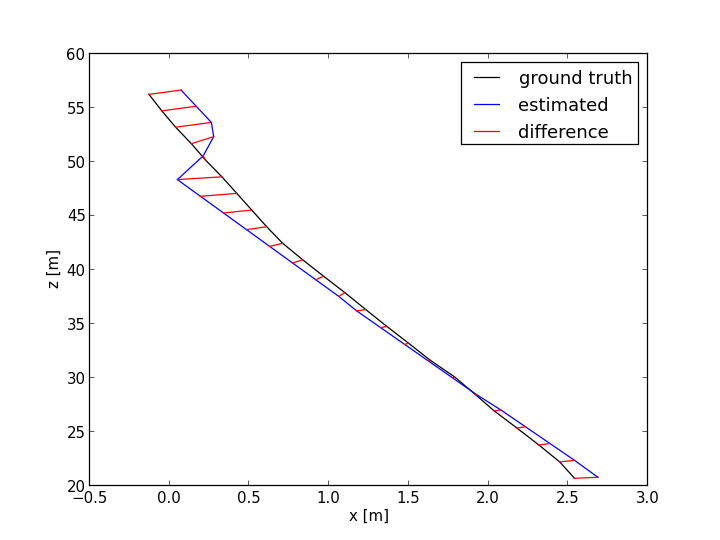} & \includegraphics[width=33mm]{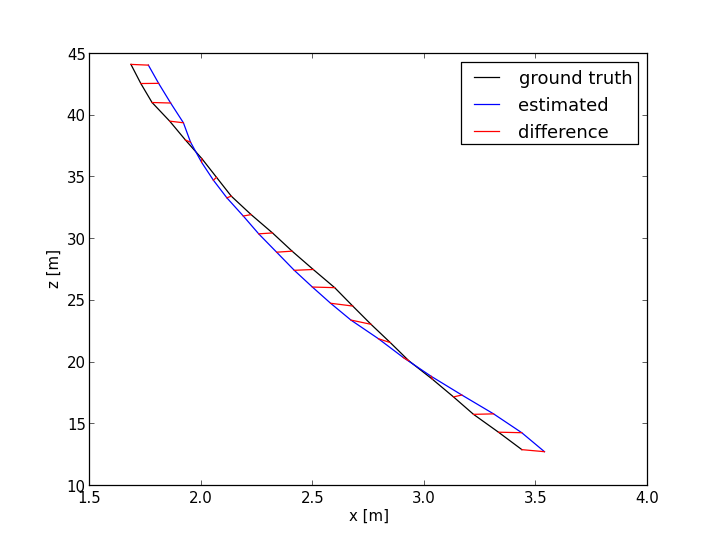}  \\
\begin{sideways}\bf \centering   \ \ \ \ \ \  OURS\end{sideways} & \includegraphics[width=33mm]{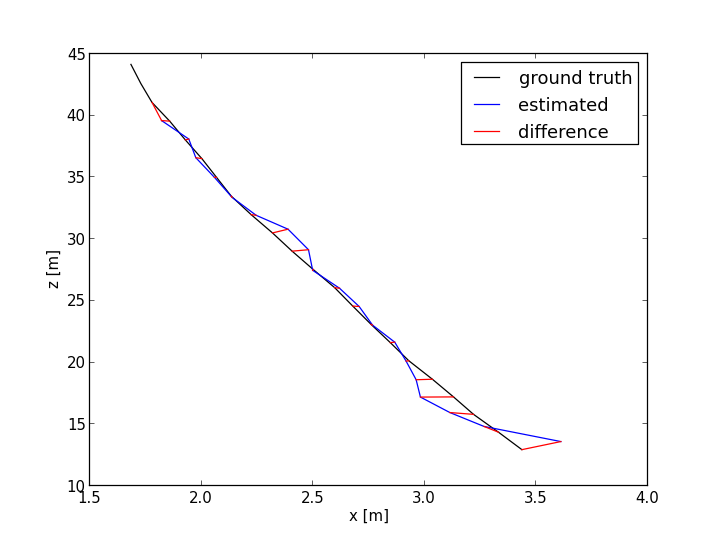} & \includegraphics[width=33mm]{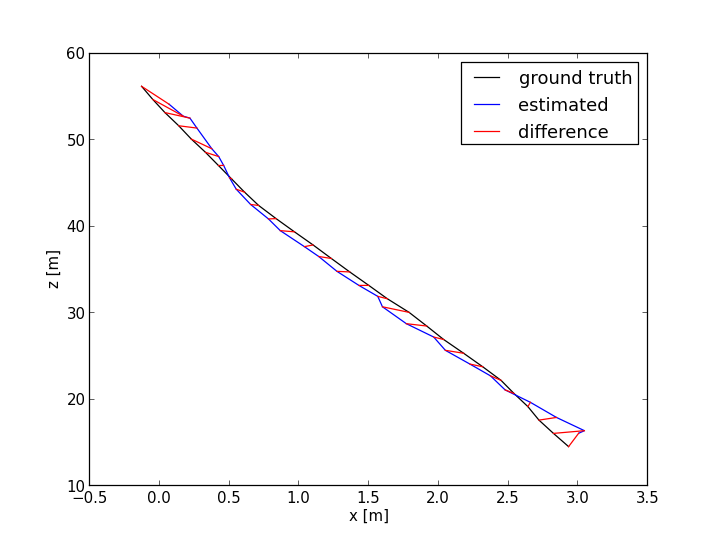} \\
\end{tabular}
\end{center}
\label{fig:plot_30}

\caption{
Comparison plots for 2 moving cars in the KITTI 2 sequence. The black plot represents the ground truth trajectory of the moving car in the world frame. Blue plot represents the estimated trajectory of the moving car. Red lines represent the error in the estimate with respect to ground truth for the trajectories. The error comparison is computed between OUR method and VISO2.
}
\end{figure}

\section{EXPERIMENTAL RESULTS}
\label{sec:exp}
We have used the KITTI tracking dataset for evaluation of the algorithm , as the ground truth localization of moving objects per camera frame is available. It consists of several sequences collected by a car-mounted camera in urban,residential and highway environments, making it a varied and challenging real world dataset. We have taken two sequences consisting of 30 images and 212 images for evaluating our algorithm. The first sequence contains two cars which are over-taking the current car and the second sequence is a highway sequence of multiple moving cars.  We chose these 2 sequences as these sequences pose challenge to motion segmentation algorithm as the moving cars lie in the subspace as the camera. These sequences also has a mix of multiple cars visible for short duration and whole sequence tracked cars which allows us to test our robustness of localization and trajectory reconstruction on both short and long sequences.  

\subsection{\bf Trajectory Evaluation}
 We compare the estimated trajectories of the moving objects to the extended Kalman filter based object tracking VISO2 (Stereo). VISO2 S(Stereo) has reported error of $2.44 \%$ on the KITTI odometry dataset, making it a good baseline algorithm to compare with. As proposed by Sturm et al.~\cite{Sturm12iros}, we compare the two sequences based on ATE for root mean square error (RMSE), mean, median and ARE. We use their evaluation algorithm which aligns the 2 trajectories using SVD. We show the three statistics as mean and median are robust to outliers while RMSE shows the exact deviation from the ground truth. We also evaluate RE which is the relative error $\%$ of RMSE: $ (ATE_{V} - ATE_{O})/ ATE_{V}$ and signifies the error change relative to VISO2. The trajectory for each moving object is computed using the KITTI tracking dataset's Ground truth. The location of each moving object in individual frame is transformed to the world reference frame using the odometry of the VISO2.

 KITTI 1 Sequence is a $212$ image sequence with in total $13$ moving cars. We have showed results for 5 moving cars (due to space constraints). The car in front of the camera is tracked for 212 images while others are 5-10 images long. As seen from Table {\ref{tab:KITTI1_ate} we can clearly observe that VISO 2 accumulates drift leading to higher RMSE and Median error over the long sequence. Our approach shows considerable robustness to drift and has average reduction error of 36.688 \%. Car 2-5 show on small sequences average error reduction of 42.5 \% which is due to the smoothness constraints.

  KITTI 2 Sequence is a $28$ image sequence with $2$ cars overtaking the camera, this poses challenge to motion segmentation leading to noisy initial estimates. Our approach here too does better than VISO2 as shown in Table~{\ref{tab:KITTI2_ate} with average error reduction of $43.57 \%$. This shows our method is able to handle both long and short sequences. Fig {\ref{fig:plot_30} shows the comparison of the trajectories relative to VISO2.

  The KITTI 2 sequence is an good example of localization error of the robot. The motion of the cars lie in the flow vector direction leading to error propagation into the camera localization. Our formulation incorporates the motion of the moving objects into the formulation causing improvement in the localization of the camera concurrently more accurate localization of landmarks and moving objects. Using our current formulation, we propose an improvement in the odometry and moving object localization when the ransac based SLAM systems fail. We can attribute this to the joint formulation of the odometry and moving object trajectories. The prior from the moving object adds to the localization of the camera.

\section{CONCLUSIONS}

This paper presents an approach for accurate localization of moving objects in a highly dynamic environment. This is an incremental real time algorithm and can be used in both indoor and outdoor environments. The algorithm solves the full multibody VSLAM optimization algorithm in real time. We have proposed a new algorithm for moving object localization. We show an improvement in the camera trajectory computation compared to the standard camera trajectory computations. An extensive evaluation of trajectories for long sequences has been compared. We propose a novel method of evaluation of moving object trajectories. The accurate localization of the moving objects is useful in ADAS systems.

We plan on releasing the GT trajectories for the moving objects and the evaluation script for cross comparison. We plan on improving the motion segmentation using additional semantic constraints for better localization of the moving objects. Incorporating additional constraints into the Factor graph will be exploited. We are also investigating the trajectory planning for the robot navigation in dynamic environments.

\addtolength{\textheight}{-12cm}   


\end{document}